\documentclass{article}

\usepackage{microtype}
\usepackage{graphicx}
\usepackage{subfigure}
\usepackage{booktabs} 
\usepackage{makecell}
\usepackage{subfigure}
\usepackage{enumitem}
\usepackage{graphicx}
\usepackage{listings}
\usepackage{multirow}
\usepackage{verbatim}
\usepackage{upgreek}

\usepackage{color}

\definecolor{gray}{rgb}{0.4,0.4,0.4}
\definecolor{dark_green}{rgb}{0.1,0.5,0.1}
\definecolor{dark_blue}{rgb}{0,0,0.7}
\usepackage{hyperref}

\usepackage[accepted]{icml2018}

\newcommand{\RLLib}[0]{{RLlib}}
\newcommand{\Ray}[0]{{Ray}}
\newcommand{\ray}[0]{{ray}}
\newcommand{\Title}[0]{{RLlib: Abstractions for Distributed Reinforcement Learning}}
\newcommand{\rlcitations}[0]{\cite{intelcoach, rllab,
hafner2017agents, baselines, pytorch-a2c-ppo-acktr,
schaarschmidt2017tensorforce}}

\newif\ifcomments

\commentstrue

\definecolor{purple}{RGB}{128,0,128}
\definecolor{indigo}{RGB}{75,0,130}

\ifcomments
\newcommand{\ion}[1]{\textbf{\textcolor{blue}{Ion: #1}}}

\newcommand{\rkn}[1]{\textbf{\textcolor{purple}{Robert: #1}}}
\newcommand{\ekl}[1]{\textbf{\textcolor{purple}{Eric: #1}}}
\newcommand{\rliaw}[1]{\textbf{\textcolor{magenta}{Richard: #1}}}
\newcommand{\pcm}[1]{\textbf{\textcolor{purple}{Philipp: #1}}}
\newcommand{\royf}[1]{\textbf{\textcolor{red}{Roy: #1}}}
\else
\newcommand{\ion}[1]{\textbf{\textcolor{blue}{}}}
\newcommand{\rkn}[1]{\textbf{\textcolor{purple}{}}}
\newcommand{\ekl}[1]{\textbf{\textcolor{purple}{}}}
\newcommand{\rliaw}[1]{\textbf{\textcolor{magenta}{}}}
\newcommand{\pcm}[1]{\textbf{\textcolor{purple}{}}}
\newcommand{\royf}[1]{\textbf{\textcolor{red}{}}}
\fi

\icmltitlerunning{\Title{}}

\begin{document}

\twocolumn[
\icmltitle{\Title{}}



\icmlsetsymbol{equal}{*}

\begin{icmlauthorlist}
\icmlauthor{Eric Liang}{equal,berkeley}
\icmlauthor{Richard Liaw}{equal,berkeley}
\icmlauthor{Philipp Moritz}{berkeley}
\icmlauthor{Robert Nishihara}{berkeley}
\icmlauthor{Roy Fox}{berkeley}
\icmlauthor{Ken Goldberg}{berkeley}
\icmlauthor{Joseph E. Gonzalez}{berkeley}
\icmlauthor{Michael I. Jordan}{berkeley}
\icmlauthor{Ion Stoica}{berkeley}
\end{icmlauthorlist}

\icmlaffiliation{berkeley}{University of California, Berkeley}

\icmlcorrespondingauthor{Eric Liang}{ericliang@berkeley.edu}

\icmlkeywords{Machine Learning, ICML}

\vskip 0.3in
]



\printAffiliationsAndNotice{\icmlEqualContribution} 

\begin{abstract}
Reinforcement learning (RL) algorithms involve the deep nesting of highly
irregular computation patterns, each of which typically exhibits opportunities
for distributed computation.
We argue for distributing RL components in a composable way by adapting algorithms
for top-down hierarchical control, thereby encapsulating parallelism and resource
requirements within short-running compute tasks.
We demonstrate the benefits of this principle through \RLLib{}: a library that
provides scalable software primitives for RL.
These primitives enable a broad range of algorithms to be implemented with
high performance, scalability, and substantial code reuse.
\RLLib{} is available as part of the open source Ray project \footnote{RLlib documentation can be found at \href{http://rllib.io}{http://rllib.io}}.

\end{abstract}

\section{Introduction}
\label{sec:intro}

Advances in parallel computing and composition through symbolic differentiation
have been fundamental to the recent success of deep learning. Today, there are a wide
range of deep learning frameworks \cite{pytorch, abadi2016tensorflow,
mxnet-learningsys, jia2014caffe} that enable rapid innovation in neural network
design and facilitate training at the scale necessary for progress in the field.


In contrast, while the reinforcement learning community enjoys the advances in
systems and abstractions for deep learning,
there has been comparatively less progress in the design of systems and abstractions
that directly target reinforcement learning.
Nonetheless, many of the challenges in reinforcement learning stem from the need
to scale learning and simulation while also integrating a rapidly increasing range
of algorithms and models. As a consequence, there is a fundamental need for
composable parallel primitives to support research in reinforcement learning.




In the absence of a single dominant computational pattern (e.g., tensor algebra)
or fundamental rules of composition (e.g., symbolic differentiation), the design
and implementation of reinforcement learning algorithms can often be cumbersome,
requiring RL researchers to directly reason about complex nested parallelism.
Unlike typical operators in deep learning frameworks,
individual components may require parallelism across a cluster (e.g., for rollouts), leverage
neural networks implemented by deep learning frameworks, recursively invoke
other components (e.g., model-based subtasks), or interface with black-box third-party simulators. In essence, the heterogeneous and distributed nature of many of these components poses a key
challenge to reasoning about their parallel composition.
Meanwhile, the main algorithms that connect these components are rapidly
evolving and expose opportunities for parallelism at varying levels. Finally, RL
algorithms manipulate substantial amounts of state (e.g., replay buffers and
model parameters) that must be managed across multiple levels of parallelism and
physical devices.

\begin{figure*}[h]
  \center
  \centering
  \begin{subfigure}[Deep Learning]{
      \includegraphics[width=3cm]{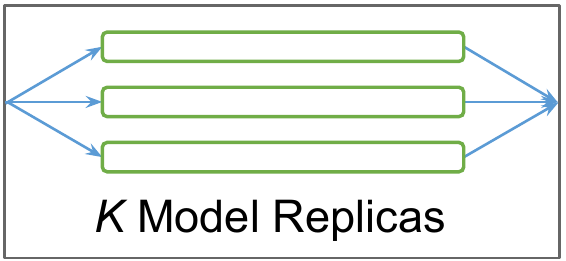}
  }
  \end{subfigure}
  \hspace{1cm}
  \begin{subfigure}[Reinforcement Learning]{
    \includegraphics[width=10.2cm]{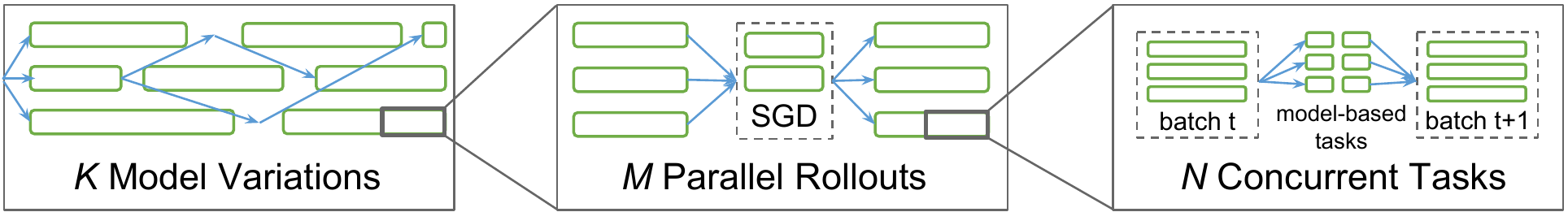}
  }
  \end{subfigure}

    \caption{
      In contrast with deep learning, RL algorithms leverage parallelism
      at multiple levels and physical devices. Here, we show an RL algorithm composing derivative-free
      optimization, policy evaluation, gradient-based optimization, and model-based planning (Table \ref{table:components}). }
  \label{fig:i2a}

\vspace{-.3cm}
\end{figure*}


The substantial recent progress in RL algorithms and applications has resulted in a
large and growing number of RL libraries \rlcitations{}. While some of these
are highly scalable, few enable the composition of components at scale. In large
part, this is due to the fact that many of the frameworks used by these
libraries rely on communication between long-running program replicas for
distributed execution; e.g., MPI \cite{mpi}, Distributed TensorFlow
\cite{abadi2016tensorflow}, and parameter servers \cite{li2014scaling}). As this
programming model ignores component boundaries, it does not naturally
encapsulate parallelism and resource requirements within individual components.\footnote{By
{\em encapsulation}, we mean that individual components specify their own
internal parallelism and resources requirements and can be used by other components
that have no knowledge of these requirements.}
As a result, reusing these distributed components requires the insertion of
appropriate control points in the program, a burdensome and error-prone process
(Section \ref{sec:proposal}).
The absence of usable encapsulation hinders code reuse and leads to error prone reimplementation of mathematically complex and often highly stochastic algorithms.
Even worse, in the distributed setting, often large parts of the distributed communication and execution must also be reimplemented with each new RL algorithm.

We believe that the ability to build scalable RL algorithms by
composing and reusing existing components and implementations is essential for
the rapid development and progress of the field.\footnote{We note that
composability {\em without} scalability can trivially be achieved with a
single-threaded library and that all of the difficulty lies in achieving these
two objectives simultaneously.} Toward this end, we argue for structuring distributed RL components around the principles of logically centralized program control and parallelism encapsulation \cite{graefe1993encapsulation, pan2010composing}.
We built \RLLib{} using these principles, and as a result were not only able to implement a broad range of state-of-the-art RL algorithms, but also to pull out scalable primitives that can be used to easily compose new algorithms.

\subsection{Irregularity of RL training workloads}

\begin{figure*}[h]
  \center
  \centering
  \begin{subfigure}[Distributed Control]{
    \label{fig:distributed_control}
    \includegraphics[width=2.7cm]{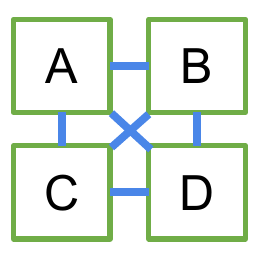}
  }
  \end{subfigure}
  \hspace{0.5cm}
  \begin{subfigure}[Logically Centralized Control]{
    \label{fig:centralized_control}
    \includegraphics[width=4.2cm]{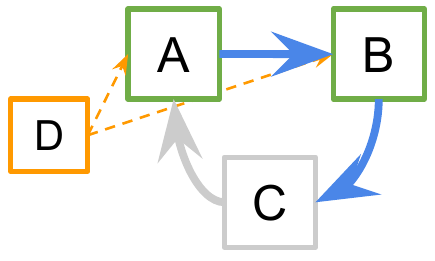}
  }
  \end{subfigure}
  \hspace{0.5cm}
  \begin{subfigure}[Hierarchical Control]{
    \label{fig:hierarchical_control}
    \includegraphics[width=7.8cm]{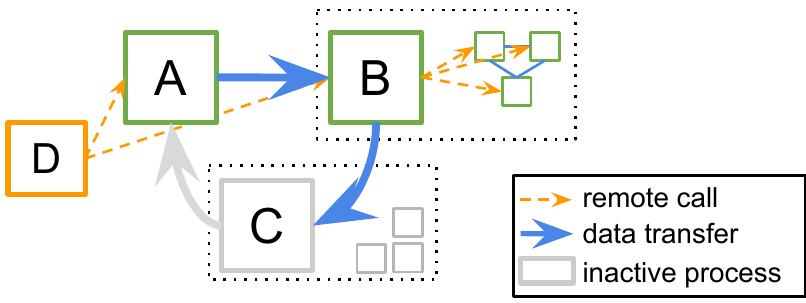}
  }
  \end{subfigure}

    \caption{
      Most RL algorithms today are written in a fully distributed style (a) where replicated processes independently
      compute and coordinate with each other according to their roles (if any). We propose a hierarchical control model (c),
      which extends (b) to support nesting in RL and hyperparameter tuning workloads, simplifying and unifying the
      programming models used for implementation. }
  \label{fig:control_models}

\end{figure*}

Modern RL algorithms are highly irregular
in the computation patterns they create (Table \ref{table:rl_irregularity}), pushing the boundaries of computation models
supported by popular distribution frameworks. This irregularity occurs at several levels:
\begin{enumerate}
  \vspace{-.2cm}
\item The duration and resource requirements of tasks differ by orders of magnitude depending on the
algorithm; e.g., A3C \cite{mnih2016asynchronous} updates may take milliseconds, but other algorithms like PPO \cite{schulman2017proximal}
batch rollouts into much larger granularities.
  \vspace{-.2cm}
\item Communication patterns vary, from synchronous to asynchronous gradient-based optimization, to having several types of asynchronous tasks in high-throughput off-policy algorithms such as Ape-X and IMPALA \cite{horgan2018distributed, espeholt2018impala}.
  \vspace{-.2cm}
\item Nested computations are generated by model-based hybrid algorithms (Table \ref{table:components}), hyperparameter tuning in conjunction with RL or DL training, or the combination of derivative-free and gradient-based optimization within a single algorithm \cite{silver2017mastering}.
  \vspace{-.2cm}
\item RL algorithms often need to maintain and update substantial amounts of state including policy parameters, replay buffers, and even external simulators.

\end{enumerate}

\begin{table}[h]
  \vspace{-.3cm}
  \caption{RL spans a broad range of computational demand.}
  \vspace{.1cm}
  \centering
  \small
  \begin{tabular}{lll}
    \toprule
      \hspace{-.2cm}Dimension & \hspace{-.2cm}DQN/Laptop & \hspace{-.2cm}IMPALA+PBT/Cluster\hspace{-.5cm}\\
    \midrule
      \hspace{-.2cm}Task Duration & \hspace{-.2cm}$\sim$1ms &  \hspace{-.2cm}minutes\hspace{-.5cm}\\
      \hspace{-.2cm}Task Compute & \hspace{-.2cm}1 CPU &  \hspace{-.2cm}several CPUs and GPUs\hspace{-.5cm}\\
      \hspace{-.2cm}Total Compute & \hspace{-.2cm}1 CPU &  \hspace{-.2cm}hundreds of CPUs and GPUs\hspace{-.5cm}\\
      \hspace{-.2cm}Nesting Depth & \hspace{-.2cm}1 level &  \hspace{-.2cm}3+ levels\hspace{-.5cm}\\
      \hspace{-.2cm}Process Memory & \hspace{-.2cm}megabytes &  \hspace{-.2cm}hundreds of gigabytes\hspace{-.5cm}\\
      \hspace{-.2cm}Execution & \hspace{-.2cm}synchronous &  \hspace{-.2cm}async. and highly concurrent\hspace{-.5cm}\\
    \bottomrule
  \end{tabular}
  \vspace{-.4cm}

  \label{table:rl_irregularity}
\end{table}

As a consequence, the developers have no choice but to use a hodgepodge of frameworks to implement their algorithms, including parameter servers, collective communication primitives in MPI-like frameworks, task queues, etc.
For more complex algorithms, it is common to build custom distributed systems in which processes
independently compute and coordinate among themselves with no central control (Figure \ref{fig:distributed_control}).
While this approach can achieve high performance, the cost to develop and evaluate is large, not only due to the need
to implement and debug distributed programs, but because composing these algorithms further complicates
their implementation (Figure \ref{fig:code_compare}). Moreover, today's computation frameworks (e.g., Spark \cite{zaharia2010spark}, MPI) typically
assume regular computation patterns and have difficulty when sub-tasks have varying durations,
resource requirements, or nesting.

\subsection{Logically centralized control for distributed RL}

It is desirable for a single programming model to capture all the requirements of RL training. This can be done without eschewing high-level frameworks that structure the computation.
Our key insight is that for each distributed RL algorithm, an equivalent algorithm can be written that exhibits
logically centralized program control (Figure \ref{fig:centralized_control}). That is, instead of having independently executing processes
(\textbf{A}, \textbf{B}, \textbf{C}, \textbf{D} in Figure \ref{fig:distributed_control}) coordinate among themselves (e.g., through RPCs, shared memory,
parameter servers, or collective communication), a single \textit{driver program} (\textbf{D} in Figure \ref{fig:centralized_control} and \ref{fig:hierarchical_control}) can delegate algorithm sub-tasks
to other processes to execute in parallel. In this
paradigm, the worker processes \textbf{A}, \textbf{B}, and \textbf{C} passively hold state (e.g., policy or simulator state) but execute no computations until called by \textbf{D}. To support nested computations,
we propose extending the centralized control model with
\textit{hierarchical delegation of control}
 (Figure \ref{fig:hierarchical_control}), which allows the worker processes (e.g., \textbf{B}, \textbf{C}) to further delegate work (e.g., simulations, gradient computation) to sub-workers of their own when executing tasks.

Building on such a logically centralized and hierarchical control model has several important advantages.
First, the equivalent algorithm is often
easier to implement in practice, since the distributed control logic is entirely encapsulated in a single
process rather than multiple processes executing concurrently. Second, the separation of algorithm
components into sub-routines (e.g., do rollouts, compute
gradients with respect to some policy loss), enables code reuse across different execution patterns.
Sub-tasks that have different resource requirements (e.g., CPUs vs GPUs) can be placed on different machines,
reducing compute costs as we show in Section \ref{sec:evaluation}.
Finally, distributed algorithms written in this
model can be seamlessly nested within each other, satisfying the parallelism
encapsulation principle.

Logically centralized control models can be highly performant, our proposed hierarchical variant even more so. This is
because the bulk of data transfer (blue arrows in Figure \ref{fig:control_models}) between processes
happens out of band of the driver, not passing through any central bottleneck. In fact many highly scalable distributed systems
\cite{zaharia2010spark, chang2008bigtable, dean2008mapreduce} leverage centralized control in their design. Within
a single differentiable tensor graph, frameworks like TensorFlow also implement logically centralized
scheduling of tensor computations onto available physical devices. Our proposal extends this principle into the broader ML systems design space.

The contributions of this paper are as follows.
\begin{enumerate}
\item We propose a general and composable hierarchical programming model for RL training (Section \ref{sec:proposal}).
\item We describe \RLLib{}, our highly scalable RL library, and how it builds on the proposed model to provide scalable abstractions for a broad range of RL algorithms, enabling rapid development (Section \ref{sec:interfaces}).
\item We discuss how performance is achieved within the proposed model (Section \ref{sec:requirements}), and show that \RLLib{} meets or exceeds state-of-the-art performance for a wide variety of RL workloads (Section \ref{sec:evaluation}).
\end{enumerate}

\section{Hierarchical Parallel Task Model}
\label{sec:proposal}

As highlighted in Figure \ref{fig:code_compare}, parallelization of entire programs
using frameworks like MPI \cite{mpi} and Distributed Tensorflow
\cite{abadi2016tensorflow} typically require explicit algorithm modifications to
insert points of coordination when trying to compose two programs or components together.
This limits the ability to rapidly prototype novel distributed RL applications.
Though the example in Figure \ref{fig:code_compare} is simple,
new hyperparameter tuning algorithms for long-running training tasks; e.g., HyperBand, Population Based Training (PBT) \cite{li2016hyperband, jaderberg2017population} increasingly demand fine-grained control over training.

We propose building RL libraries with hierarchical and logically centralized control on top
of flexible task-based programming models like \Ray{} \cite{moritz2017ray}.
Task-based systems
allow subroutines to be scheduled and executed asynchronously on worker processes,
on a fine-grained basis, and for results to be retrieved or passed between processes.

\subsection{Relation to existing distributed ML abstractions}

Though typically formulated for distributed control, abstractions such as parameter servers
and collective communication operations can also be used within a logically centralized control model. As an example,
\RLLib{} uses allreduce and parameter-servers in some of its policy optimizers (Figure \ref{fig:sgd_sync_vs_async}),
and we evaluate their performance in Section \ref{sec:evaluation}.


\subsection{\Ray{} implementation of hierarchical control}
\label{sec:ray_api}

We note that, within a single machine, the proposed
programming model can be implemented simply with thread-pools and shared memory, though it
is desirable for the underlying framework to scale to larger clusters if needed.

We chose to build \RLLib{} on top of the \Ray{} framework, which
allows Python tasks to be distributed across large clusters. \Ray{}'s distributed scheduler
is a natural fit for the hierarchical control model, as nested computation can be implemented in Ray
with no central task scheduling bottleneck.

To implement a logically centralized control model, it is first necessary to have a mechanism
to launch new processes and schedule tasks on them. \Ray{} meets this requirement with
\textit{\Ray{} actors}, which are Python classes that may be created in the cluster and
accept remote method calls (i.e., tasks). \Ray{} permits these actors to in turn launch more actors
and schedule tasks on those actors as part of a method call, satisfying our need for hierarchical
delegation as well.

For performance, \Ray{} provides standard communication primitives such as \texttt{\small{aggregate}}
and \texttt{\small{broadcast}}, and critically enables the \textit{zero-copy} sharing of large data objects
through a shared memory object store. As shown in Section \ref{sec:evaluation}, this enables
the performance of \RLLib{} algorithms. We further discuss framework performance in Section \ref{sec:requirements}.

\begin{figure}[bt]
  \centering
  \begin{subfigure}[Distributed Control]{
      \includegraphics[width=3.9cm]{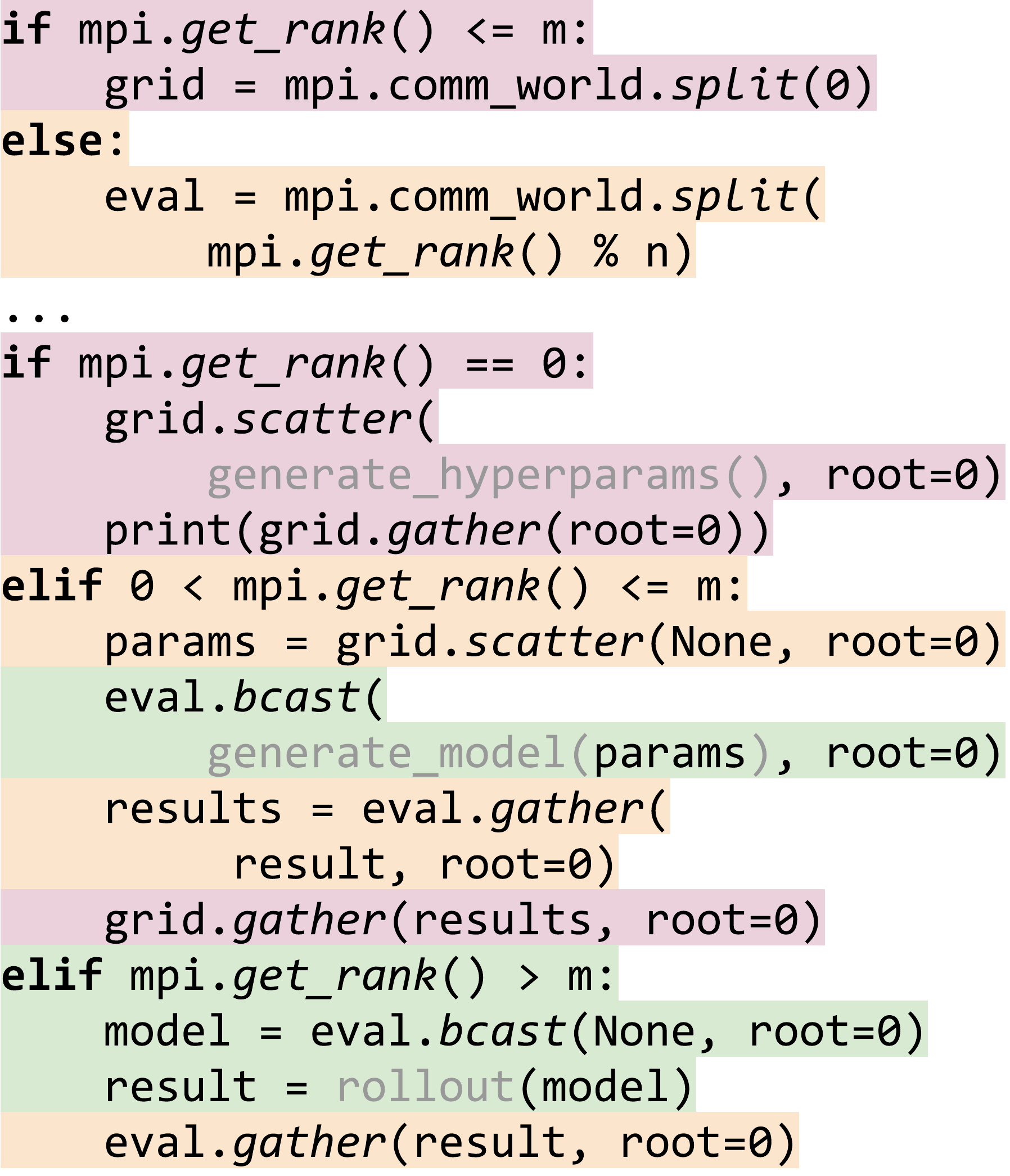}
  }
  \end{subfigure}
  \begin{subfigure}[Hierarchical Control]{
      \includegraphics[width=3.4cm]{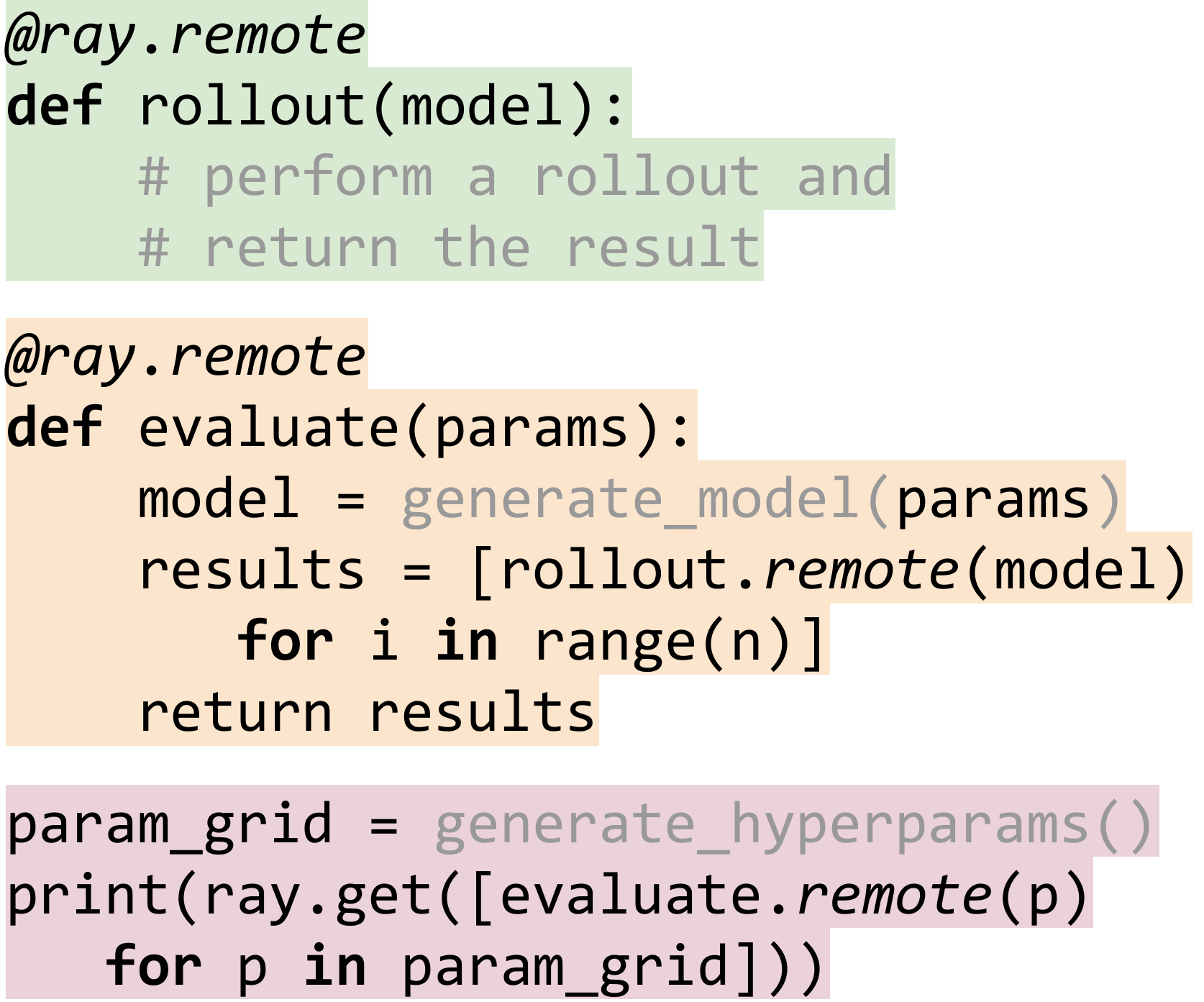}
  }
  \end{subfigure}

  \caption{
  Composing a distributed hyperparameter search with a function
  that also requires distributed computation involves {\em complex nested
  parallel computation patterns}. With MPI (a), a new program must
  be written from scratch that mixes elements of both. With hierarchical
  control (b), components can remain unchanged and simply be invoked as remote tasks.
  }
  \vspace{-.2cm}

  \label{fig:code_compare}

\end{figure}

\section{Abstractions for Reinforcement Learning}
\label{sec:interfaces}

\begin{figure*}[ht!]
  \vspace{-0.3cm}
  \centering
  \hspace{-0.5cm}
  \begin{subfigure}[Allreduce]{
      \includegraphics[width=3.7cm]{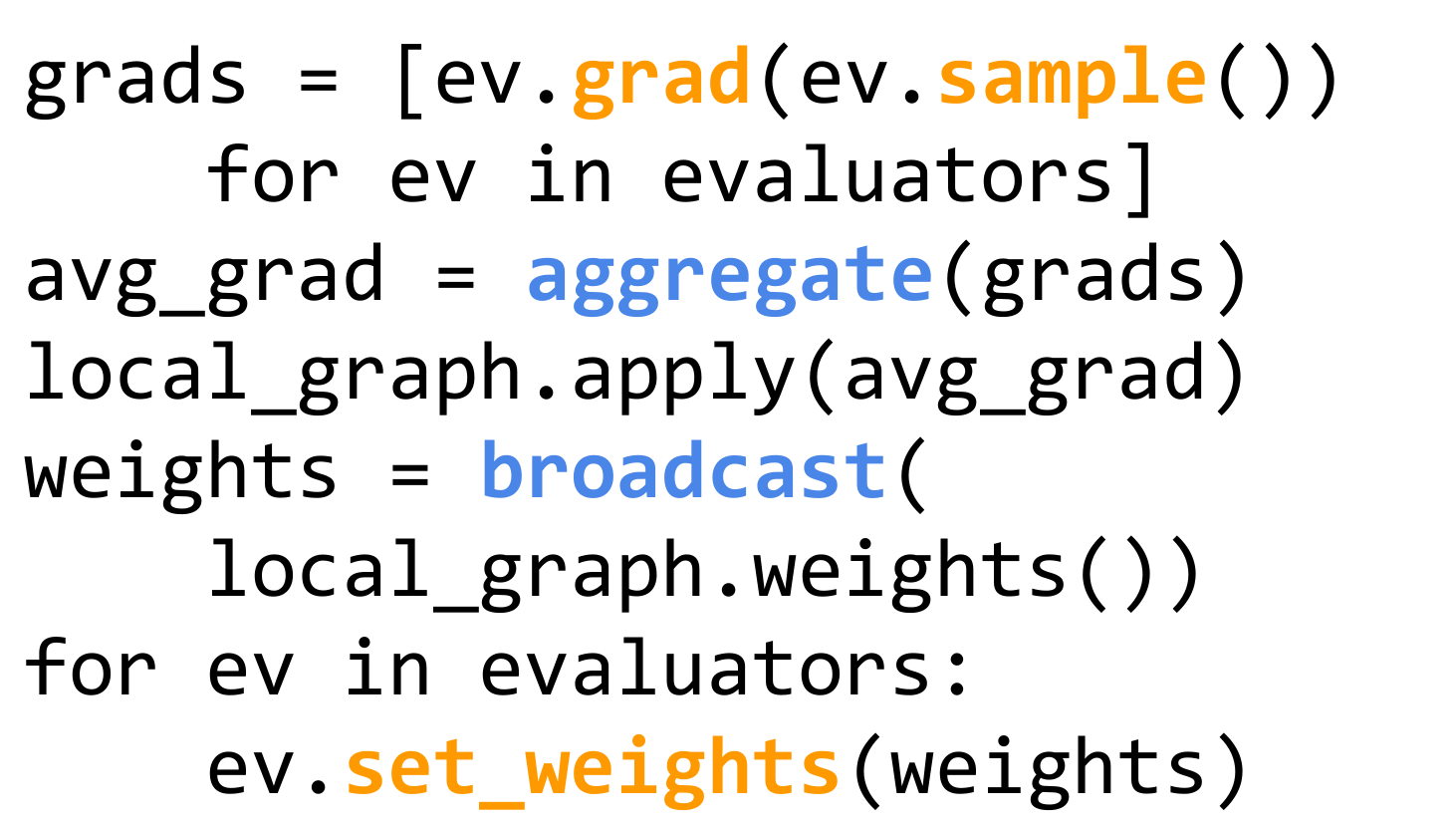}
  }
  \hspace{-0.5cm}
  \end{subfigure}
  \begin{subfigure}[Local Multi-GPU]{
      \includegraphics[width=4.6cm]{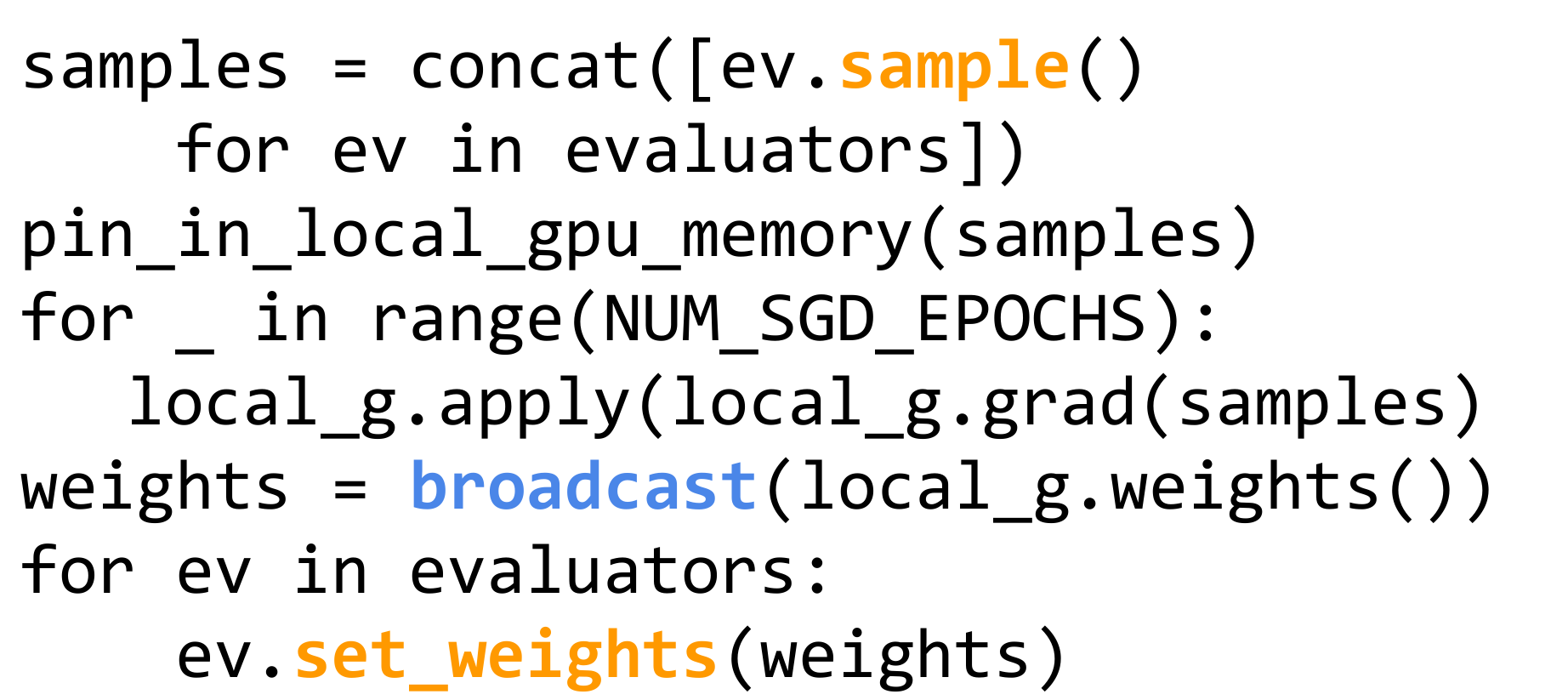}
      \label{fig:multi_gpu}
  }
  \hspace{-0.7cm}
  \end{subfigure}
  \begin{subfigure}[Asynchronous]{
      \label{fig:async_optimizer}
      \includegraphics[width=4.4cm]{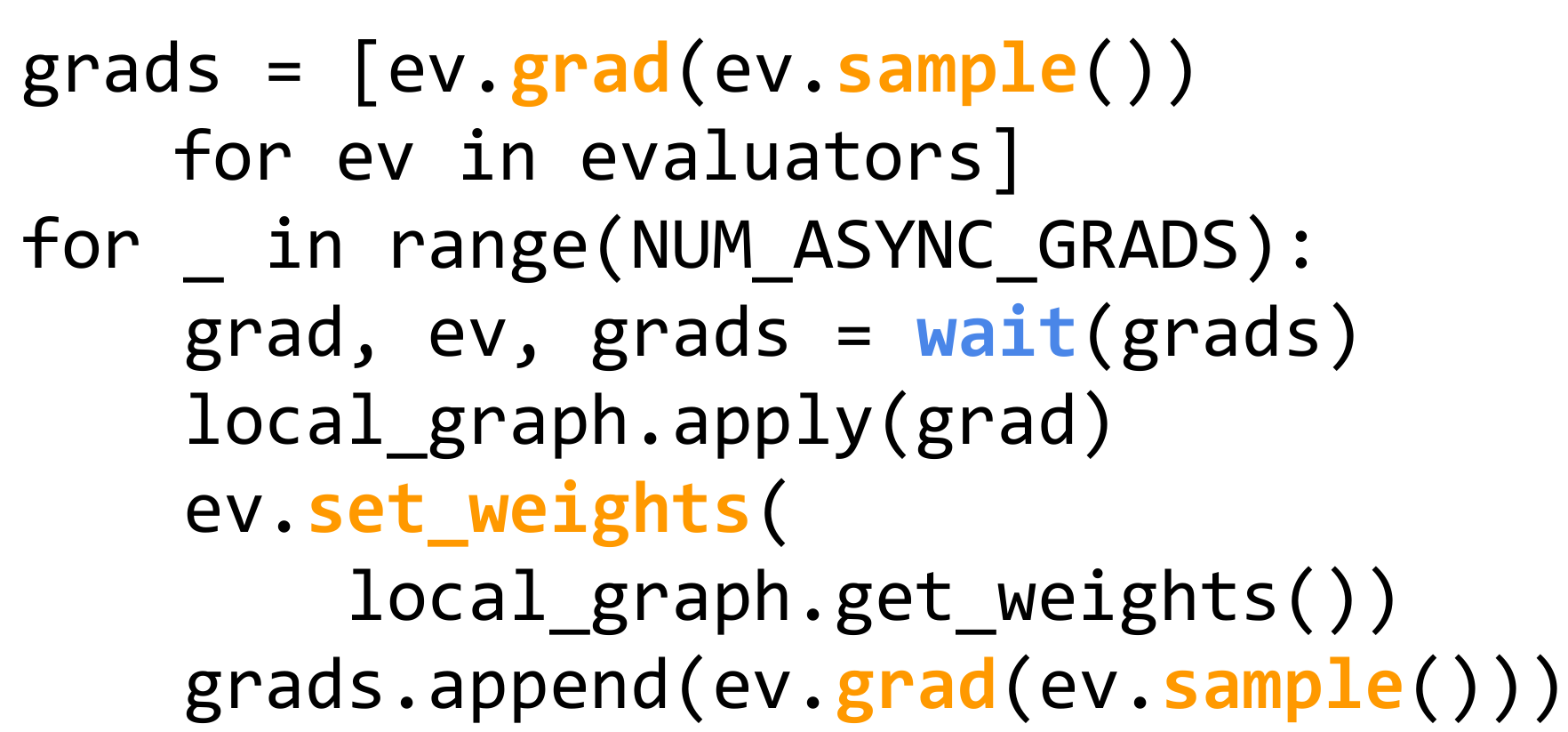}
      \label{fig:async_opt}
  }
  \hspace{-0.4cm}
  \end{subfigure}
  \begin{subfigure}[Sharded Param-server]{
      \includegraphics[width=4.6cm]{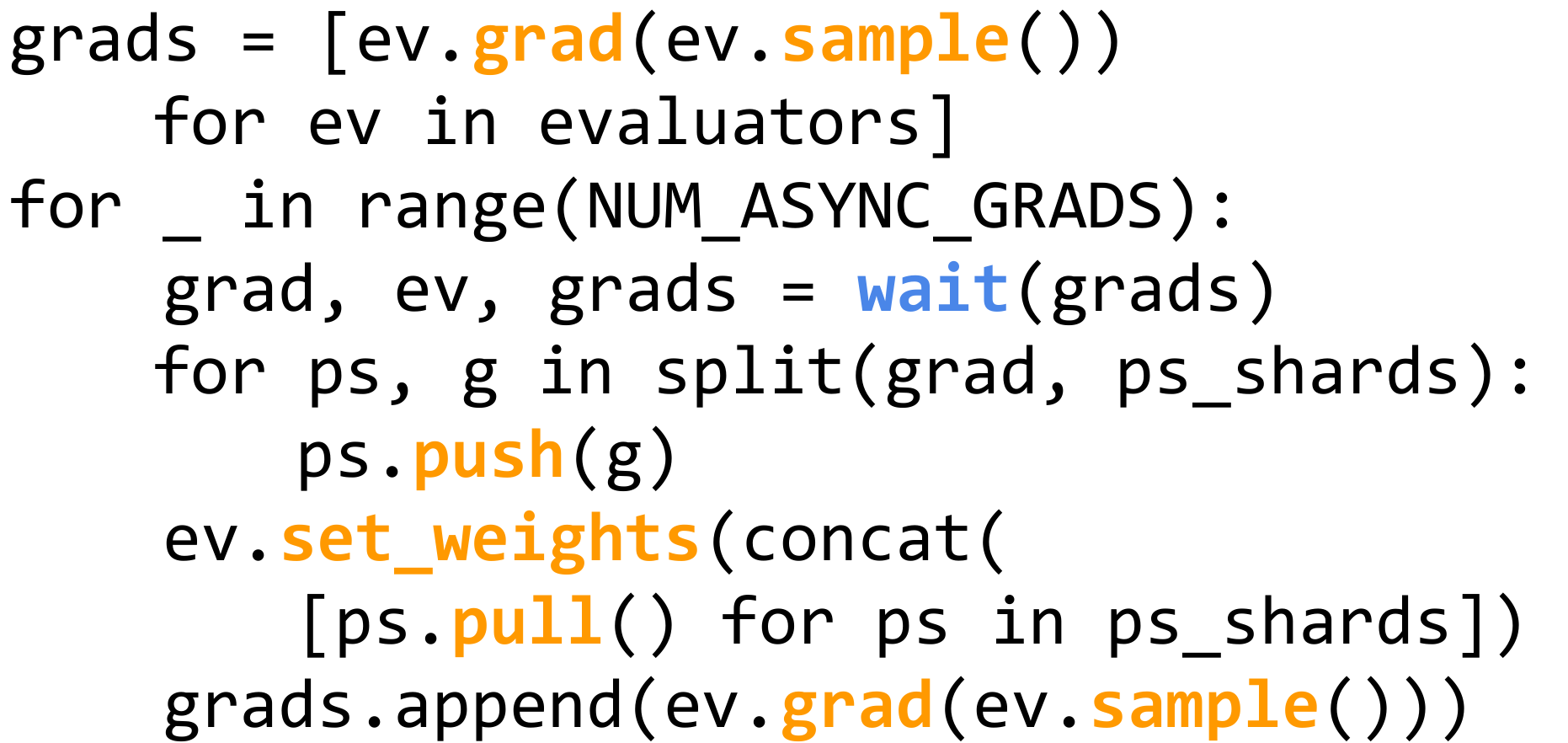}
      \label{fig:sharded_ps}
  }
  \hspace{-0.5cm}
  \end{subfigure}

    \caption{Pseudocode for four \RLLib{} policy optimizer step methods. Each step() operates over a local policy graph and array of remote evaluator replicas. \Ray{} remote
    calls are highlighted in orange; other \Ray{} primitives in blue (Section
    \ref{sec:requirements}). \textit{Apply} is shorthand for updating weights.
    Minibatch code and helper functions omitted. The param server optimizer in \RLLib{} also implements pipelining not shown here.}
  \label{fig:sgd_sync_vs_async}
\end{figure*}

To leverage \RLLib{} for distributed execution, algorithms must declare their policy $\pi$, experience postprocessor $\rho$, and loss $L$. These can be specified in any deep learning framework, including TensorFlow and PyTorch. \RLLib{} provides \textit{policy evaluators} and \textit{policy optimizers} that implement strategies for distributed policy evaluation and training.



\subsection{Defining the Policy Graph}

RLlib's abstractions are as follows. The developer specifies a policy model $\pi$ that maps the current observation $o_t$ and (optional) RNN hidden state $h_t$ to an action $a_t$ and the next RNN state $h_{t+1}$. Any number of user-defined values $y^i_t$ (e.g., value predictions, TD error) can also be returned:
\vspace{-.05cm}
\begin{equation}
\pi_{\theta}(o_t, h_t) \Rightarrow (a_t, h_{t+1}, y^1_{t}\ldots{}y^N_{t})
\end{equation}
\vspace{-.05cm}
Most algorithms will also specify a trajectory postprocessor $\rho$ that transforms a batch $X_{t,K}$ of $K$ $\{(o_t, h_t, a_t, h_{t+1}, y^1_{t}\ldots{}y^N_{t}, r_t, o_{t+1})\}$ tuples starting at $t$. Here $r_t$ and $o_{t+1}$ are the reward and new observation after taking an action. Example uses include advantage estimation \cite{schulman2015high} and goal relabeling \cite{andrychowicz2017hindsight}. To also support multi-agent environments, experience batches $X_{t,K}^p$ from the $P$ other agents in the environment are also made accessible:
\vspace{-.05cm}
\begin{equation}
\rho_{\theta}(X_{t,K}, X_{t,K}^1\ldots{}X_{t,K}^P) \Rightarrow X_{post}
\end{equation}
\vspace{-.05cm}
Gradient-based algorithms define a combined loss $L$ that can be descended to improve the policy and auxiliary networks:
\vspace{-.05cm}
\begin{equation}
L(\theta{}; X) \Rightarrow loss
\end{equation}
\vspace{-.05cm}
Finally, the developer can also specify any number of utility functions $u^i$ to be called as needed during training to, e.g., return training statistics $s$, update target networks, or adjust annealing schedules:
\vspace{-.05cm}
\begin{equation}
u^1\ldots{}u^M(\theta) \Rightarrow (s, \theta_{update})
\end{equation}
\vspace{-.05cm}
To interface with RLlib, these algorithm functions should be defined in a \textit{policy graph} class with the following methods:

\vspace{.1cm}
\begin{lstlisting}
abstract class rllib.PolicyGraph:
	def act(self, obs, h): action, h, y*
	def postprocess(self, batch, b*): batch
	def gradients(self, batch): grads
	def get_weights; def set_weights;
	def u*(self, args*)
\end{lstlisting}
\vspace{-.3cm}

\subsection{Policy Evaluation}
For collecting experiences, \RLLib{} provides a \texttt{\small{PolicyEvaluator}} class that wraps a policy graph and environment to add a method to \texttt{\small{sample()}} experience batches. Policy evaluator instances can be created as Ray remote actors and \textit{replicated} across a cluster for parallelism. To make their usage concrete, consider a minimal TensorFlow policy gradients implementation that extends the \texttt{\small{rllib.TFPolicyGraph}} helper template:

\vspace{.1cm}
\begin{lstlisting}
class PolicyGradient(TFPolicyGraph):
  def __init__(self, obs_space, act_space):
    self.obs, self.advantages = ...
    pi = FullyConnectedNetwork(self.obs)
    dist = rllib.action_dist(act_space, pi)
    self.act = dist.sample()
    self.loss = -tf.reduce_mean(
      dist.logp(self.act) * self.advantages)
  def postprocess(self, batch):
    return rllib.compute_advantages(batch)
\end{lstlisting}
\vspace{-.1cm}

From this policy graph definition, the developer can create a number of policy evaluator replicas \texttt{\small{ev}} and call \texttt{\small{ev.sample.remote()}} on each to collect experiences in parallel from environments. \RLLib{} supports OpenAI Gym \cite{gym}, user-defined environments, and also batched simulators such as ELF \cite{elf}:

\vspace{.1cm}
\begin{lstlisting}
evaluators = [rllib.PolicyEvaluator.remote(
    env=SomeEnv, graph=PolicyGradient)
  for _ in range(10)]
print(ray.get([
    ev.sample.remote() for ev in evaluators]))
\end{lstlisting}
\vspace{-.1cm}

\subsection{Policy Optimization}
\label{subsec:pluggable_sgd}

\RLLib{} separates the implementation of algorithms into the declaration of the algorithm-specific \textit{policy graph} and the choice of an algorithm-independent \textit{policy optimizer}. The policy optimizer is responsible for the performance-critical tasks of distributed sampling, parameter updates, and managing replay buffers. To distribute the computation, the optimizer operates over a set of policy evaluator replicas.

To complete the example, the developer chooses a policy optimizer and creates it with references to existing evaluators. The async optimizer uses the evaluator actors to compute gradients in parallel on many CPUs (Figure \ref{fig:async_optimizer}). Each \texttt{\small{optimizer.step()}} runs a round of remote tasks to improve the model. Between steps, policy graph replicas can be queried directly, e.g., to print out training statistics:

\vspace{.1cm}
\begin{lstlisting}
optimizer = rllib.AsyncPolicyOptimizer(
    graph=PolicyGradient, workers=evaluators)
while True:
    optimizer.step()
    print(optimizer.foreach_policy(
        lambda p: p.get_train_stats()))
\end{lstlisting}
\vspace{-.1cm}

Policy optimizers extend the well-known gradient-descent optimizer abstraction to the RL domain. A typical gradient-descent optimizer implements $step(L(\theta), X, \theta) \Rightarrow \theta_{opt}$. \RLLib{}'s policy optimizers instead operate over the local policy graph $G$ and a set of remote evaluator replicas, i.e., ${step(G, ev_1\dots{}ev_n, \theta) \Rightarrow \theta_{opt}}$, capturing the sampling phase of RL as part of optimization (i.e., calling \texttt{\small{sample()}} on policy evaluators to produce new simulation data).

The policy optimizer abstraction has the following advantages. By separating execution strategy from policy and loss definitions, specialized optimizers can be swapped in to take advantage of available hardware or algorithm features without needing to change the rest of the algorithm. The policy graph class encapsulates interaction with the deep learning framework, allowing algorithm authors to avoid mixing distributed systems code with numerical computations, and enabling optimizer implementations to be improved and reused across different deep learning frameworks.

As shown in Figure \ref{fig:sgd_sync_vs_async}, by leveraging centralized control, policy optimizers succinctly capture a broad range of choices in RL optimization: synchronous vs asynchronous, allreduce vs parameter server, and use of GPUs vs CPUs.
\RLLib{}'s policy optimizers provide
performance comparable to optimized parameter server (Figure \ref{subfig:ps}) and MPI-based implementations (Section \ref{sec:evaluation}).
Pulling out this optimizer abstraction is easy in a logically centralized control model since each policy optimizer has full control over the distributed computation it implements.

\begin{table*}
\small

  \caption{
  \RLLib{}'s policy optimizers and evaluators capture common components (Evaluation, Replay, Gradient-based Optimizer)
  within a logically centralized control model, and leverages \Ray{}'s hierarchical task model
  to support other distributed components.
  }
  \vspace{0.3cm}
  \centering
  \centering
  \begin{tabular}{lcccc}
\toprule
   Algorithm Family & Policy Evaluation & Replay Buffer & Gradient-Based Optimizer & Other Distributed Components\\
\midrule
    DQNs
                      & X & X & X &  \\
    Policy Gradient
                      & X &   & X &  \\
    Off-policy PG
                      & X & X & X &  \\
    Model-Based/Hybrid
                      & X &   & X & Model-Based Planning\\
    Multi-Agent
                      & X & X & X &  \\
    Evolutionary Methods
                      & X &   &   &  Derivative-Free Optimization\\
    AlphaGo
                      & X & X & X & MCTS, Derivative-Free Optimization\\

  \end{tabular}

  \label{table:components}
  \vspace{-0.3cm}

\end{table*}

\subsection{Completeness and Generality of Abstractions}
\label{sec:completeness}

\begin{figure}
  \centering
  \centering
  \begin{subfigure}[Sharded Param. Server]{
      \label{subfig:ps}
      \includegraphics[width=3.8cm]{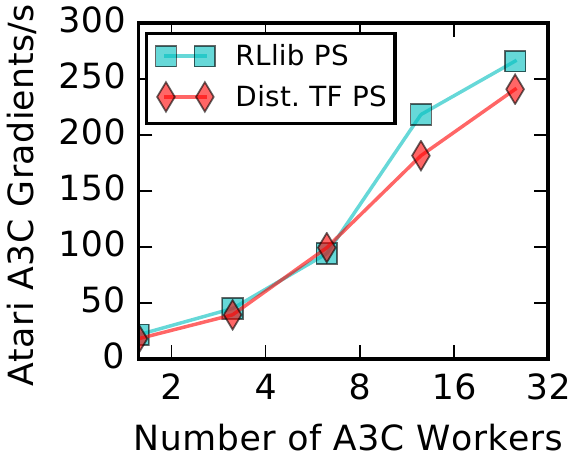}
  }
  \end{subfigure}
  \begin{subfigure}[Ape-X in \RLLib{}]{
      \label{subfig:apex}
    \includegraphics[width=3.8cm]{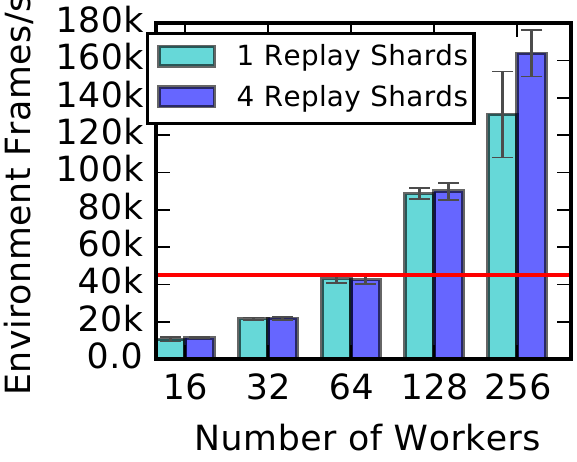}
  }
  \end{subfigure}

    \vspace{-.2cm}
    \caption{\RLLib{}'s centrally controlled policy optimizers match or exceed the performance of implementations in specialized systems. The \RLLib{} parameter server optimizer using 8 internal shards is competitive with a Distributed TensorFlow implementation tested in similar conditions. \RLLib{}'s Ape-X policy optimizer scales to 160k frames per second with 256 workers at a frameskip of 4, more than matching a reference throughput of $\sim$45k fps at 256 workers, demonstrating that a single-threaded Python controller can efficiently scale to high throughputs.}
    \vspace{-.1cm}

\end{figure}

We demonstrate the completeness of \RLLib{}'s abstractions by formulating the algorithm
families listed in Table \ref{table:components} within the API. When applicable, we also
describe the concrete implementation in \RLLib{}:

\textbf{DQNs}: DQNs use $y^1$ for storing TD error, implement n-step return calculation in $\rho_{\theta}$, and the Q loss in $L$. Target updates are implemented in $u^1$, and setting the exploration $\epsilon$ in $u^2$.

\textit{DQN implementation}: To support experience replay, \RLLib{}'s DQN uses a policy optimizer that saves collected samples in an embedded replay buffer. The user can alternatively use an asynchronous optimizer (Figure \ref{fig:async_opt}). The target network is updated by calls to $u^1$ between optimizer steps.

\textit{Ape-X implementation}: Ape-X \cite{horgan2018distributed} is a variation of DQN that leverages distributed experience prioritization to scale to many hundreds of cores. To adapt our DQN implementation, we created policy evaluators with a distribution of $\epsilon$ values, and wrote a new high-throughput policy optimizer ($\sim$200 lines) that pipelines the sampling and transfer of data between replay buffer actors using Ray primitives. Our implementation scales nearly linearly up to 160k environment frames per second with 256 workers (Figure \ref{subfig:apex}), and the optimizer can compute gradients for $\sim$8.5k 80$\times$80$\times$4 observations/s on a V100 GPU.

\textbf{Policy Gradient / Off-policy PG}: These algorithms store value predictions in $y^1$, implement advantage estimation using $\rho_{\theta}$, and combine actor and critic losses in $L$.

\textit{PPO implementation}: Since PPO's loss function permits multiple SGD passes over sample data, when there is sufficient GPU memory \RLLib{} chooses a GPU-optimized policy
optimizer (Figure \ref{fig:multi_gpu}) that pins data into local GPU memory. In each
iteration, the optimizer collects samples from evaluator replicas, performs multi-GPU optimization
locally, and then broadcasts the new model weights.

\textit{A3C implementation}: \RLLib{}'s A3C can use either the asynchronous (Figure \ref{fig:async_opt}) or sharded parameter server policy optimizer (\ref{fig:sharded_ps}). These optimizers collect gradients from the policy evaluators to update the canonical copy of $\theta$.

\textit{DDPG implementation}: \RLLib{}'s DDPG uses the same replay policy optimizer as DQN. $L$ includes both actor and critic losses. The user can also choose to use the Ape-X policy optimizer with DDPG.

\textbf{Model-based / Hybrid}: Model-based RL algorithms extend $\pi_{\theta}(o_t, h_t)$ to make decisions based on model rollouts, which can be parallelized using Ray. To update their environment models, the model loss can either be bundled with $L$, or the model trained separately (i.e., in parallel using Ray primitives) and its weights periodically updated via $u^1$.

\textbf{Multi-Agent}: Policy evaluators can run multiple policies at once in the same environment, producing batches of experience for each agent. Many multi-agent algorithms use a centralized critic or value function, which we support by allowing $\rho_{\theta}$ to collate experiences from multiple agents.

\textbf{Evolutionary Methods}: Derivative-free methods can be supported through non-gradient-based policy optimizers.

\begin{figure}[ht!]
  \vspace{-.2cm}
  \centering
  \centering
  \begin{subfigure}[Evolution Strategies]{
      \includegraphics[width=3.7cm]{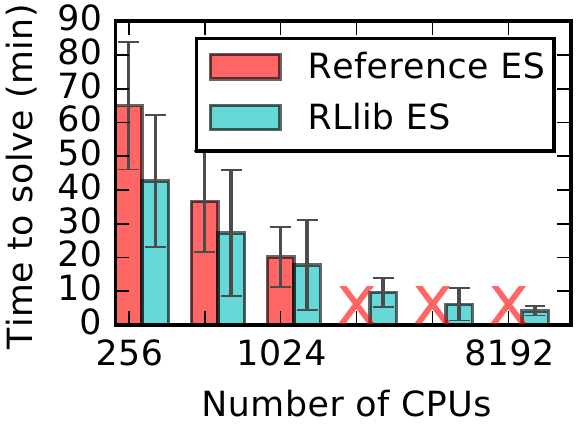}
      \label{fig:es_scaling}
  }
  \end{subfigure}
  \begin{subfigure}[PPO]{
    \includegraphics[width=4.0cm]{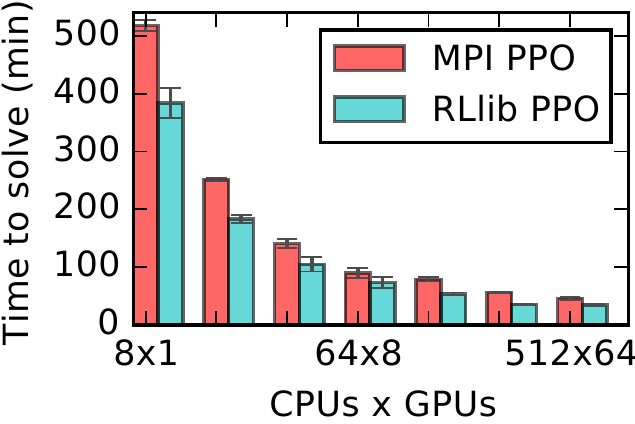}
  }
  \end{subfigure}

    \vspace{-.4cm}
    \caption{The time required to achieve a reward of 6000 on the Humanoid-v1
    task. \RLLib{} implementations of ES and PPO outperform highly optimized
    reference optimizations.}
    \vspace{-.3cm}

\end{figure}

\textit{Evolution Strategies (ES) implementation}: ES is a derivative-free optimization algorithm that scales well to clusters with thousands of CPUs. We were able to port a
single-threaded implementation of ES to \RLLib{} with only a few
changes, and further scale it with an aggregation tree of actors (Figure \ref{fig:es_scaling}), suggesting that the hierarchical control model is both flexible and easy to adapt algorithms for.

\textit{PPO-ES experiment}: We studied a hybrid algorithm that runs PPO
updates in the inner loop of an ES optimization step that randomly perturbs
the PPO models. The implementation took only $\sim$50 lines of code and did not require changes to PPO, showing the value of encapsulating parallelism.
In our experiments, PPO-ES converged faster and to a higher reward than PPO on the Walker2d-v1 task. A similarly modified A3C-ES
implementation solved PongDeterministic-v4 in 30\% less time.

\begin{figure*}
  \centering
  \begin{subfigure}[Ape-X]{
      \includegraphics[width=5cm]{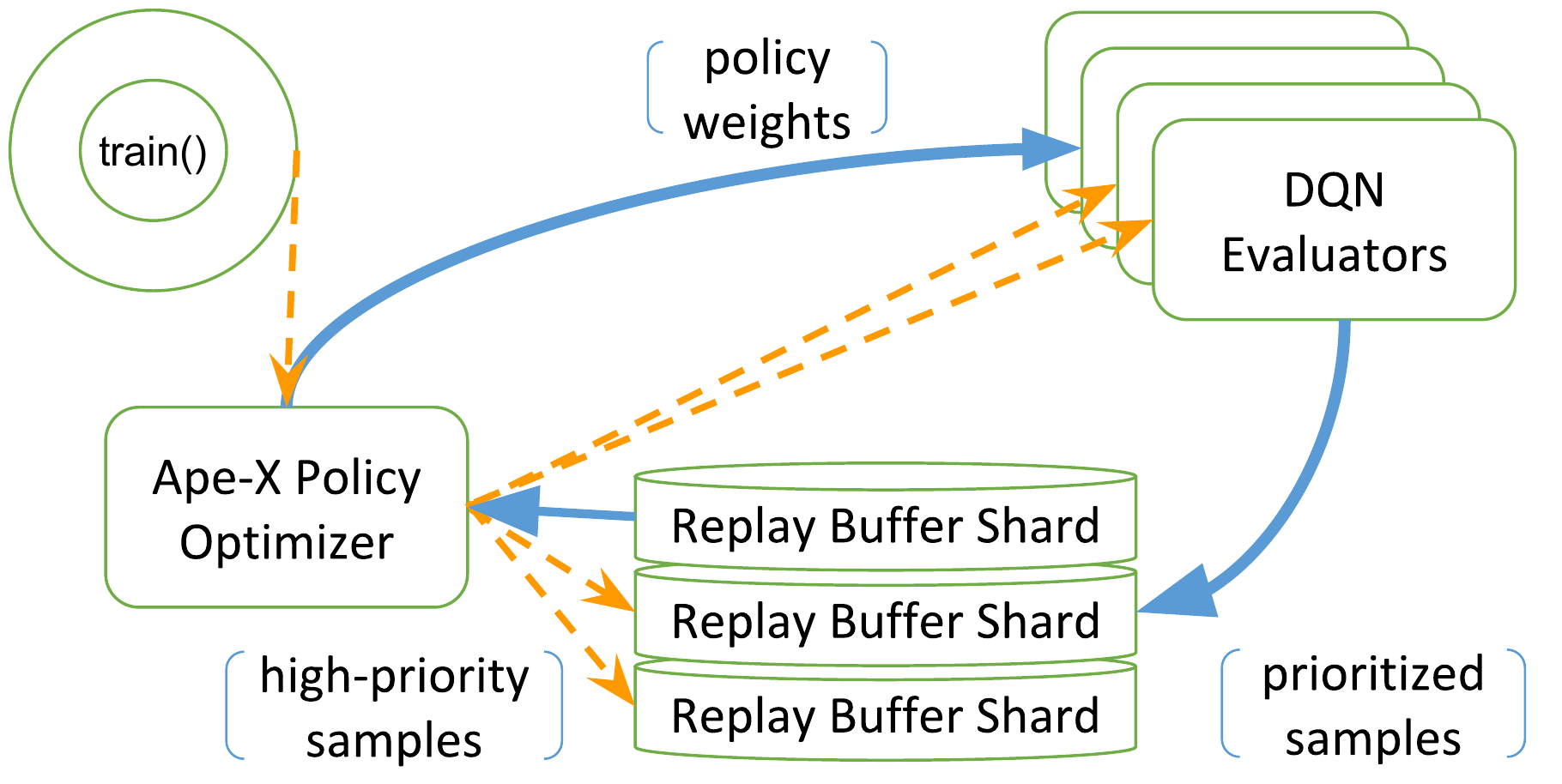}
  }
  \end{subfigure}
  \begin{subfigure}[PPO-ES]{
      \includegraphics[width=5cm]{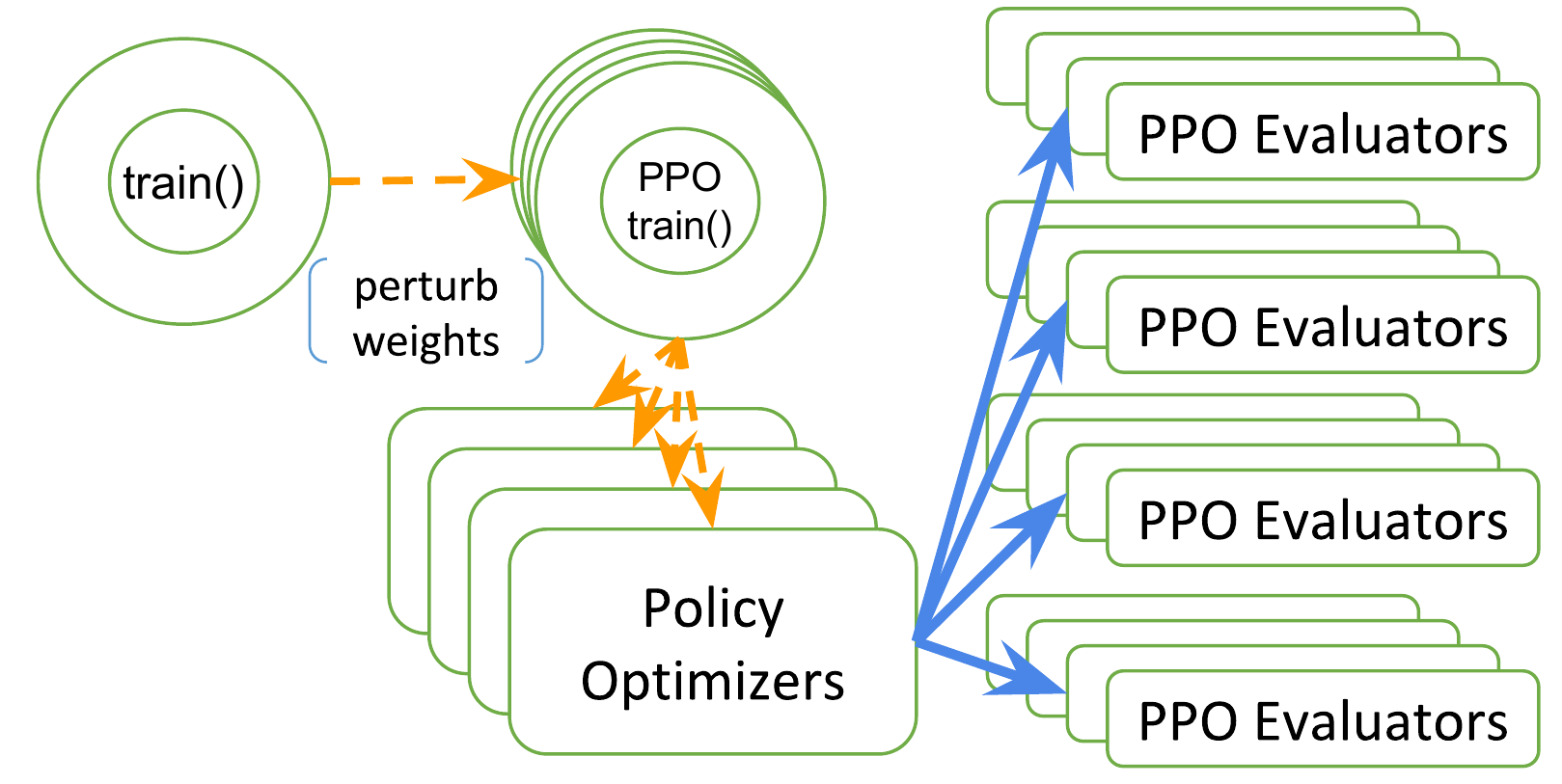}
  }
  \end{subfigure}
  \begin{subfigure}[AlphaGo Zero]{
      \includegraphics[width=5cm]{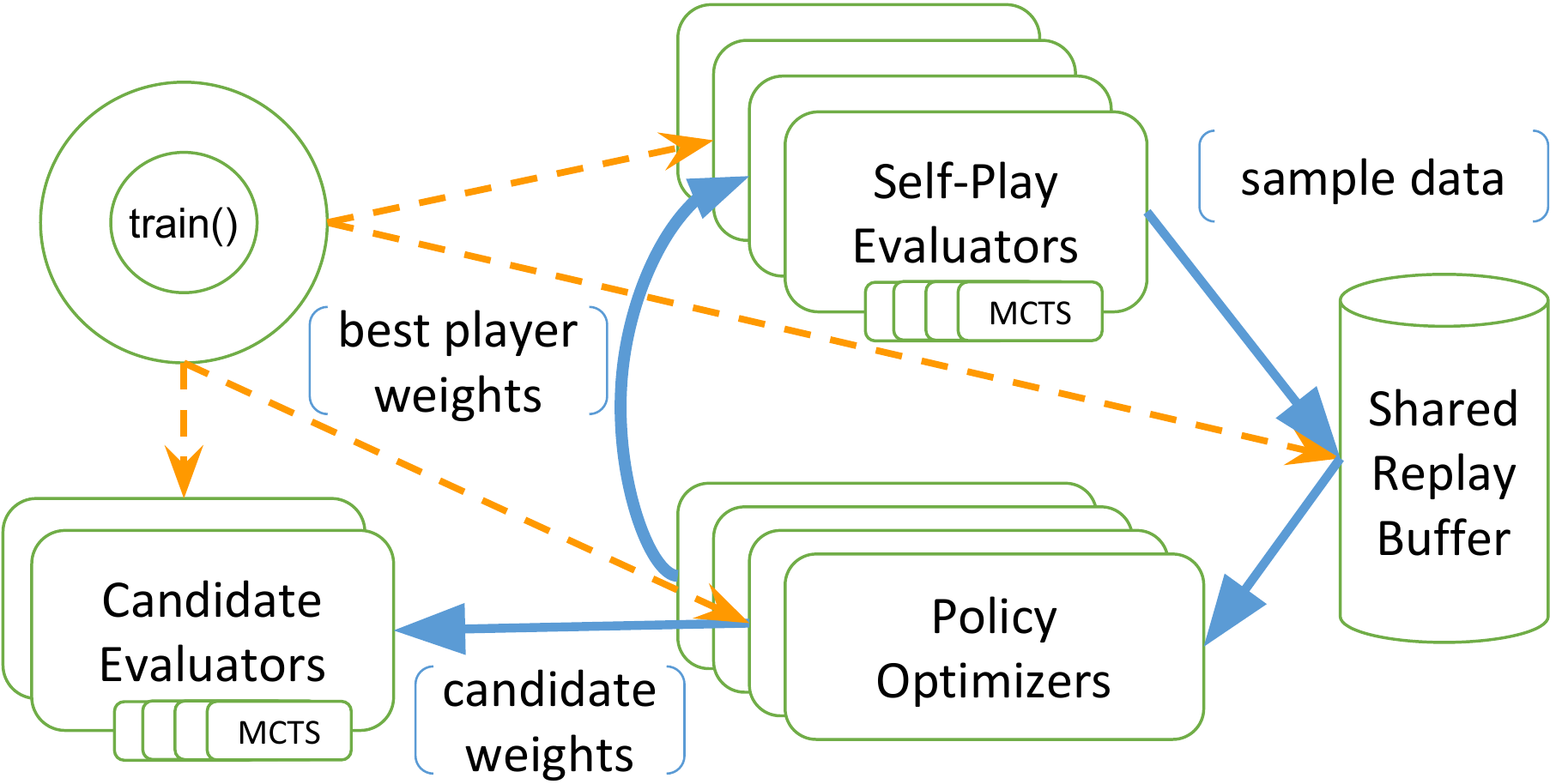}
  }
  \end{subfigure}

  \vspace{-.2cm}
  \caption{Complex RL architectures are easily captured within \RLLib{}'s hierarchical control model.
  Here blue lines denote data
  transfers, orange lines lighter overhead method calls. Each train() call
  encompasses a batch of remote calls between components.}

  \label{fig:complex_arch}
  \vspace{-.1cm}
\end{figure*}

\textbf{AlphaGo}: We sketch how to scalably implement the AlphaGo Zero algorithm using a combination of \Ray{} and \RLLib{} abstractions. Pseudocode for the $\sim$70 line main algorithm loop is provided in the Supplementary Material.

\begin{enumerate}[wide, labelwidth=!, labelindent=8pt]

  \vspace{-.3cm}
  \item \textit{Logically centralized control of multiple distributed components}: AlphaGo Zero uses multiple distributed components: model
  optimizers, self-play evaluators, candidate model evaluators, and the shared
  replay buffer. These components are manageable as Ray
  actors under a top-level AlphaGo policy optimizer. Each optimizer step loops over actor
  statuses to process new results, routing data between actors and launching
  new actor instances.

  \vspace{-.1cm}
  \item \textit{Shared replay buffer}: AlphaGo Zero stores the experiences from
  self-play evaluator instances in a shared replay buffer. This requires
  routing game results to the
  shared buffer, which is easily done by passing the result object references
  from actor to actor.

  \vspace{-.1cm}
  \item \textit{Best player}: AlphaGo Zero tracks the current best model and
  only populates its replay buffer with self-play from that model. Candidate
  models must achieve a $\geq55\%$ victory margin to replace the best model.
  Implementing this amounts to adding an \texttt{\small{if}} block in the main
  control loop.

  \vspace{-.1cm}
  \item \textit{Monte-Carlo tree search}: MCTS (i.e., model-based planning) can be handled as a subroutine of the policy graph, and optionally parallelized as well using Ray.

\end{enumerate}

\textbf{HyperBand and Population Based Training}: \Ray{} includes
distributed implementations of hyperparameter search algorithms such as HyperBand and PBT
\cite{li2016hyperband, jaderberg2017population}. We were able to use these to evaluate \RLLib{} algorithms,
which are themselves distributed, with the addition of $\sim$15 lines of code per
algorithm.
We note that these algorithms are non-trivial to integrate when using distributed control models due to the need to modify existing code to insert points of coordination (Figure \ref{fig:code_compare}). \RLLib{}'s use of short-running tasks avoids this problem, since control decisions can be easily made between tasks.

\section{Framework Performance}
\label{sec:requirements}

In this section, we discuss properties of \Ray{} \cite{moritz2017ray} and other
optimizations critical to \RLLib{}.

\subsection{Single-node performance}

\textbf{Stateful computation:} Tasks can share mutable state with other tasks
through Ray actors. This is critical for tasks that operate on and mutate stateful objects like third-party simulators or neural network weights.

\textbf{Shared memory object store:} RL workloads involve sharing large
quantities of data (e.g., rollouts and neural network weights).
\Ray{} supports this by allowing data objects to be passed directly between workers
without any central bottleneck. In \Ray{}, workers on the same machine can
also read data objects through shared memory without copies.

\textbf{Vectorization:} \RLLib{} can batch policy evaluation to improve hardware utilization (Figure \ref{fig:sampling_scaling}), supports batched environments, and passes experience data between actors efficiently in columnar array format.

\subsection{Distributed performance}

\textbf{Lightweight tasks:} Remote call overheads in \Ray{} are on the order of
$\sim$200$\upmu$s when scheduled on the same machine. When machine resources are saturated,
tasks spill over to other nodes, increasing latencies to
around $\sim$1ms. This enables parallel algorithms to scale seamlessly to
multiple machines while preserving high single-node throughput.

\textbf{Nested parallelism:} Building RL algorithms by composing distributed
components creates multiple levels of nested parallel calls (Figure
\ref{fig:i2a}). Since components make decisions that may affect downstream
calls, the call graph is also inherently dynamic. \Ray{} supports this by allowing any Python function or class method to be invoked remotely as a
lightweight task. For example, \texttt{\small{func.\textbf{remote}()}} executes
\texttt{\small{func}} remotely and immediately returns a placeholder result which can
later be retrieved or passed to other tasks.

\textbf{Resource awareness:}
\Ray{} allows remote calls to specify resource requirements and utilizes a
resource-aware scheduler to preserve component performance.
Without this, distributed components can improperly allocate resources, causing algorithms to run inefficiently or fail.



\textbf{Fault tolerance and straggler mitigation:} Failure events become significant at scale
\cite{barroso2013datacenter}. \RLLib{} leverages \Ray{}'s built-in fault tolerance
mechanisms \cite{moritz2017ray}, reducing costs with preemptible
cloud compute instances \cite{spot, preemptible}. Similarly, stragglers can significantly impact the
performance of distributed algorithms at scale \cite{dean2013tail}. \RLLib{}
supports straggler mitigation in a generic way via the
\texttt{\small{\ray{}.\textbf{wait}()}} primitive. For example, in PPO we use this to
drop the slowest evaluator tasks, at the cost of some bias.

\textbf{Data compression:} \RLLib{} uses the LZ4 algorithm to compress experience batches.
For image observations, LZ4 reduces network traffic and memory usage by more than an order of
magnitude, at a compression rate of $\sim$1 GB/s/core.

\section{Evaluation}
\label{sec:evaluation}

\begin{figure}[ht!]
  \centering
  \includegraphics[width=8.2cm]{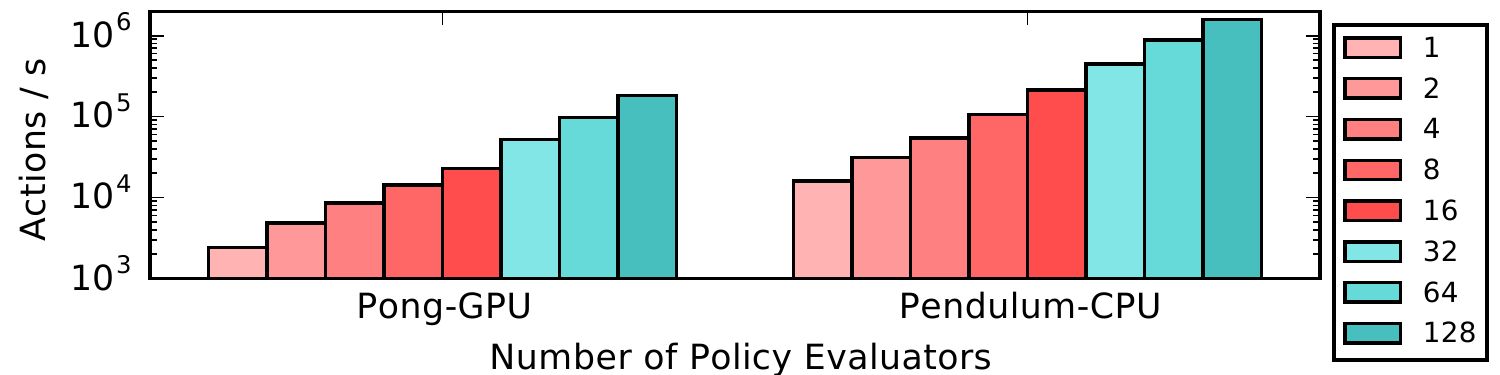}
  \vspace{-.7cm}
  \caption{Policy evaluation throughput scales nearly linearly from 1 to 128 cores. PongNoFrameskip-v4 on GPU scales from 2.4k to $\sim$200k actions/s, and Pendulum-v0 on CPU from 15k to 1.5M actions/s. We use a single p3.16xl AWS instance to evaluate from 1-16 cores, and a cluster of four p3.16xl instances from 32-128 cores, spreading actors evenly across the cluster. Evaluators compute actions for 64 agents at a time, and share the GPUs on the machine.}
  \label{fig:sampling_scaling}
\end{figure}

\textbf{Sampling efficiency:} Policy evaluation is an important building block for all RL algorithms. In Figure \ref{fig:sampling_scaling} we benchmark the scalability of gathering samples from policy evaluator actors. To avoid bottlenecks, we use four intermediate actors for aggregation. Pendulum-CPU reaches over 1.5 million actions/s running a small 64$\times$64 fully connected network as the policy. Pong-GPU nears 200k actions/s on the DQN convolutional architecture \cite{mnih2015human}.

\textbf{Large-scale tests:} We evaluate the performance of \RLLib{} on Evolution Strategies (ES), Proximal
Policy Optimization (PPO), and A3C, comparing against specialized systems built
\textit{specifically for those algorithms} \cite{openaies, baselines,
openaistarter} using Redis, OpenMPI, and Distributed TensorFlow. The same
hyperparameters were used in all experiments. We used TensorFlow
to define neural networks for the \RLLib{} algorithms evaluated.

\RLLib{}'s ES implementation scales well on the Humanoid-v1 task to 8192 cores
using AWS m4.16xl CPU instances \cite{awspricing}. With 8192 cores, we achieve a
reward of 6000 in a median time of 3.7 minutes, which is over twice as fast as the best
published result \cite{salimans2017evolution}. For PPO we evaluate on the
same Humanoid-v1 task, starting with one p2.16xl GPU instance and adding m4.16xl
instances to scale. This
cost-efficient local policy optimizer (Table \ref{table:sgd_tradeoffs}) outperformed
the reference MPI implementation that required multiple expensive GPU instances
to scale.

We ran \RLLib{}'s A3C on an x1.16xl machine and solved the PongDeterministic-v4
environment in 12 minutes using an asynchronous policy optimizer and 9 minutes using a sharded param-server optimizer, which matches the performance of a well-tuned baseline \cite{openaistarter}.

\textbf{Multi-GPU:} To better understand \RLLib{}'s advantage in the PPO
experiment, we ran benchmarks on a p2.16xl instance comparing \RLLib{}'s
local multi-GPU policy optimizer with one using an allreduce in Table
\ref{table:sgd_tradeoffs}. The fact that different strategies perform better
under different conditions suggests that policy optimizers are a useful abstraction.

\begin{table}[hb!]

  \scriptsize
  \centering
  \begin{tabular}{llll}
    \toprule
      \hspace{-.2cm}Policy Optimizer\hspace{-.2cm}&\hspace{-.2cm}Gradients computed on \hspace{-.3cm}& \hspace{-.3cm}Environment \hspace{-.3cm}& \hspace{-.3cm}SGD throughput \hspace{-.3cm}\\
    \midrule
      \multirow{4}{*}{\hspace{-.2cm}Allreduce-based\hspace{-.2cm}} & \multirow{2}{*}{\hspace{-.2cm}4 GPUs, Evaluators\hspace{-.3cm}} & \hspace{-.3cm}Humanoid-v1 & \hspace{-.3cm}330k samples/s   \hspace{-.3cm}   \\
      & & \hspace{-.3cm}Pong-v0    & \hspace{-.3cm}23k samples/s  \hspace{-.3cm}\\
        \cmidrule{2-4}
      & \multirow{2}{*}{\hspace{-.2cm}16 GPUs, Evaluators\hspace{-.3cm}}  & \hspace{-.3cm}Humanoid-v1 \hspace{-.3cm}& \hspace{-.3cm}\textbf{440k samples/s}  \hspace{-.3cm}    \\
      & & \hspace{-.3cm}Pong-v0 & \hspace{-.3cm}\textbf{100k samples/s} \hspace{-.3cm} \\
    \midrule
      \multirow{4}{*}{\hspace{-.2cm}Local Multi-GPU}\hspace{-.3cm} & \multirow{2}{*}{\hspace{-.2cm}4 GPUs, Driver\hspace{-.3cm}} & \hspace{-.3cm}Humanoid-v1 & \hspace{-.3cm}\textbf{2.1M samples/s} \hspace{-.3cm}\\
      & & \hspace{-.3cm}Pong-v0 & \hspace{-.3cm}N/A (out of mem.) \hspace{-.3cm}\\
        \cmidrule{2-4}
      & \multirow{2}{*}{\hspace{-.2cm}16 GPUs, Driver\hspace{-.3cm}} & \hspace{-.3cm}Humanoid-v1 & \hspace{-.3cm}1.7M samples/s \hspace{-.3cm}\\
      & & \hspace{-.3cm}Pong-v0 & \hspace{-.3cm}\textbf{150k samples/s} \hspace{-.3cm}\\
    \bottomrule
  \end{tabular}
  \caption{A specialized multi-GPU policy optimizer outperforms distributed
  allreduce when data can fit entirely into GPU memory. This experiment was done
  for PPO with 64 Evaluator processes. The PPO batch size was 320k, The SGD
  batch size was 32k, and we used 20 SGD passes per PPO batch.}

  \label{table:sgd_tradeoffs}
  \vspace{-.4cm}
\end{table}

\vspace{-.1cm}
\section{Related work}
\vspace{-.1cm}
\label{sec:related}

There are many reinforcement learning libraries \rlcitations{}. These often scale by creating long-running program replicas that each participate in coordinating the distributed computation as a whole, and as a result do not generalize well to complex architectures.
\RLLib{} instead uses a hierarchical control model with short-running tasks to let each component control its own distributed execution, enabling higher-level abstractions such as policy optimizers to be used for composing and scaling RL algorithms.



Outside of reinforcement learning, there has been a strong effort to explore
composition and integration between different deep learning frameworks. ONNX
\cite{microsoft_onnx}, NNVM \cite{nnvm}, and Gluon \cite{gluon} sit between
model specifications and hardware to provide cross-library optimizations. Deep
learning libraries \cite{pytorch, abadi2016tensorflow, mxnet-learningsys,
jia2014caffe} provide support for the gradient-based optimization components that appear in RL algorithms.


\vspace{-.2cm}
\section{Conclusion}
\vspace{-.1cm}
\RLLib{} is an open source library for reinforcement learning that leverages fine-grained nested parallelism to achieve state-of-the-art performance across a broad range of RL workloads. It offers both a collection
of reference algorithms and scalable abstractions for easily composing new ones.

\pagebreak

\section*{Acknowledgements}

In addition to NSF CISE Expeditions Award CCF-1730628, 
this research is supported in part by DHS Award HSHQDC-16-3-00083, 
and gifts from Alibaba, Amazon Web Services, Ant Financial, Arm,
CapitalOne, Ericsson, Facebook, Google, Huawei, Intel, Microsoft, Scotiabank, Splunk and VMware.

\bibliography{main}
\bibliographystyle{icml2018}

\end{document}